# Reinforcement Learning for Robotic Safe Control with Force Sensing

Nan Lin[1], Linrui Zhang[2], Yuxuan Chen[1], ZhenruiChen[2], Yujun Zhu[1],
Ruoxi Chen[2], Peichen Wu[1] and Xiaoping Chen[1]

*Abstract*— For the task with complicated manipulation in unstructured environments, traditional hand-coded methods are ineffective, while reinforcement learning can provide more general and useful policy. Although the reinforcement learning is able to obtain impressive results, its stability and reliability is hard to guarantee, which would cause the potential safety threats. Besides, the transfer from simulation to real-world also will lead in unpredictable situations. To enhance the safety and reliability of robots, we introduce the force and haptic perception into reinforcement learning. Force and tactual sensation play key roles in robotic dynamic control and human-robot interaction. We demonstrate that the force-based reinforcement learning method can be more adaptive to environment, especially in sim-to-real transfer. Experimental results show in object pushing task, our strategy is safer and more efficient in both simulation and real world, thus it holds prospects for a wide variety of robotic applications.

## I. INTRODUCTION

Reinforcement learning has been widely applied in a range of robotic control, from autonomous car [1], unmanned aerial vehicle [2] to dexterous anthropomorphic manipulation [3]. In simulation environments, reinforcement learning even works better than human experts. However, considering that robot would meet a complex, dynamic and unstructured environment which cannot be reproduced totally in simulation, small errors may lead to failure. This kind of sim-to-real problem pervasively exists in robotic tasks [4], [5], and improper transfer often causes safety problems. Reinforcement learning can not guarantee the safety at present, for the policy it has learned is highly nonlinear, even in simulation the accidental instance may crop up, let alone in the real-world situations with unavoidable noise and errors.

We need some additional measures to prevent dangerous behavior when the policy is out of control. Meanwhile, the methods should be integrated into the reinforcement learning framework to improve efficiency. The research of robot safety has got sufficient attention and development recently, in cooperative robots [6] and soft robots [7], even the soft-tissue injury model could be accurately established [8]. But in reinforcement learning, those force control strategies are rarely used. Here, we adopt a common force-torque sensor and touch sensors and demonstrate that in object pushing task, a simple force-based reward and corrective feedback action can improve the safety effectively, and significantly reduce the impact and abnormal behaviors. An illustration of safe manipulation is shown in Fig. 1.

*This work is supported by the National Natural Science Foundation of China under grant U1613216 and 61573333.
[1]Nan Lin, Yuxuan Chen, Yujun Zhu, Peichen Wu and Xiaoping Chen are with the School of Computer Science and Technology, University of Science and Technology of China, Hefei, 230026, China. fhln@mail.ustc.edu.cn
[2]Linrui Zhang, Zhenrui Chen and Ruoxi Chen are with the School of Information Science and Technology, University of Science and Technology of China, Hefei, 230026, China.

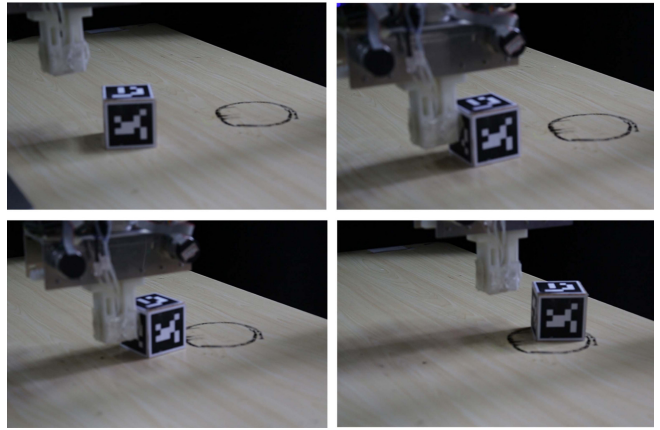

Fig. 1. The object pushing task with safe control strategy.

## II. RELATED WORK

Robotic manipulation with object is a widely studied issue in the field of robotics. With the booming of the reinforcement learning, various model-free algorithms using deep neural networks have been applied. Tasks such as pushing, relocating, and grasping have been well solved in simulation. Kovac et al. proposed a solution for pusher-watcher and it is proved to be effective [10]. Kalashnikov et al. improved the control based on closed-loop vision, which increased the success rate of grasping objects and reduced the environmental requirements [11]. Zeng et al. collaborated the pushing tasks with grasping tasks to complete more complex tasks [4]. There are also some studies [12] that focus on the transfer of pushing and other tasks in real-world environments using implicit system identification and modeling. But the need of significant amounts of data is a major difficulty of reinforcement learning at present.

Training directly on the real robots is not feasible because of safety and other issues [13]. Although there are some examples of direct training in real world [14], for multi-dimensional work space or highly nonlinear dynamics, it is more reasonable to obtain data for efficient exploration through human demonstrations. Lowrey et al. used the reinforcement learning method to train in the simulation environment and achieved good experimental results in the real environment [15]. However, sim-to-real transfer not only needs to bridge the reality gap, but also needs to better ensure the rationality and safety of the control from the engineering perspective.

Robot safety is indeed an important research area, which should be guaranteed first in human-robot interaction and collaboration. Using flexible materials and stiffness tuning, the compliance can be improved. And torque sensors or series elastic actuators [16] in joints also enhance performance, combining impedance/admittance control [17]. Other methods includes skin sensors [18] and visual-based detection [19], etc. In reinforcement learning, there have been some researches on robot safety. For example, using the Lyapunov stability theory to explicitly consider safety [20], setting the state constraints [21]–[23], or specifying risk-aversion in the reward [24]. But in the dynamic environments, those methods may lose efficacy. In robotic applications, joint angles can be directly controlled to adapt to the environment [12], or sense the 6-axis force and produce compliant behavior [25], but sometimes the robot will still be beyond control.

Here, the force-based control is introduced into reinforcement learning. More specifically, we adopt the tactile and force/torque sensors to apperceive the outside environment. The force and haptic perception not only enhances efficiency of exploration, but also detects the hazardous situations and takes safe actions to avoid harming the environment or damaging the robot. We test our algorithm both in simulation and real world, and experimental results further confirm that our strategy has a distinct advantage over the ordinary reinforcement learning methods.

## III. BACKGROUND

### A. reinforcement learning

The reinforcement learning methods applied to the robotics is a basically control problems. An agent acts in a stochastic environment by selecting actions in a sequential manner, maximizing the cumulative rewards perceived by the environment. The problem can be modeled as a Markov decision process, including a state space $S$, an action space $A$, an initial state distribution density $p_1(s_1)$, a state transition density $p(s_{t+1}|s_t, a_t)$ that satisfies the Markov property for all trajectories in the state-action space, and a reward function $R(s_t, a_t) : S \times A \to \mathbb{R}$.

At time $t$, the agent's state is $s_t$, and the agent chooses and executes the action $a_t$ according to the policy $\pi(a_t|s_t)$. Then the agent's state converts to a new state $s_t$ according to $p(s_t|s_t, a_t)$ and the agent obtains a reward $r(s_t, a_t)$. Finally we can get a cumulative reward $R(\tau) = \sum_{t=0}^{T} \gamma^t r_t$. We assume that the environment transformation and policy are probability distributions. In this case, the T-steps trajectory is:

$$P(\tau|\pi) = \rho_0(s_0) \prod_{t=0}^{T-1} P(s_{t+1}|s_t, a_t) \pi(a_t|s_t) \quad (1)$$

Expected return is $J(\pi) = \int_\tau P(\tau|\pi) R(\tau) = \mathop{\mathrm{E}}_{\tau \sim \pi}[R(\tau)]$. The goal is to find the optimal policy $\pi^*$ to maximize the expected return [26].

### B. Deep Deterministic Policy Gradient

Deep Deterministic Policy Gradient (DDPG) [27] is a model-free and off-policy reinforcement learning algorithm which uses deep neural networks for function approximation. The algorithm needs to maintain two networks, i.e., the actor and the critic. The actor acts as an approximator of the strategy $\pi : S \to A$ which infers that the action $A$ is determined according to the current state $S$ (in order to balance the exploration and exploit, a random noise $\mathcal{N}_t$ is usually added to the actual action), and critic approximates the action-value function $Q : S \times A \to \mathbb{R}$ to judge the advantage of certain action.

The training process optimizes the Actor parameters $\theta^\pi$ and the Critic parameters $\varphi^Q$ at the same time. Specifically, the definition of the critic's Loss function is

$$L = \frac{1}{N} \sum_i (y_i - Q(s_i, a_i))^2 \quad (2)$$

where

$$y_i = r_i + \gamma Q(s_{i+1}, \pi(s_{i+1})) \quad (3)$$

Based on the standard back-propagation method, $\nabla_{\varphi^Q} L$ can be computed. For the actor's policy gradient $J$ [28], the unbiased estimate of $J$ according to the Monto-Carlo method is

$$\nabla_{\theta^\pi} J(\pi) = \frac{1}{N} \sum_i \left( \nabla_a Q(s,a) |_{s=s_i, a=\pi(s_i)} \cdot \nabla_{\theta^\pi} \pi(s) |_{s=s_i} \right) \quad (4)$$

### C. Hindsight Experience Replay

The combination of Hindsight Experience Replay (HER) and DDPG has been approved to greatly improve the training speed and efficiency of sparse-reward tasks [29]. For example, a form of reward is shaped as:

$$r(s, g) = \begin{cases} 0, & \text{if } g \text{ is satisfied in } s \\ -1, & \text{otherwise} \end{cases} \quad (5)$$

In most cases in training, we get -1 as the reward, which is not conducive to the optimization of network parameters. If we have some other goals achieved, the situation could be improved greatly. HER defines a mapping $m : S \to G$ for $\tau = (s_0, a_0 ... a_{T-1}, s_T)$ in every episode. According to the new mission target $g'$, we can get the new $\tau'$. Then in the process of replay, it will have much more successful examples. A simple approach proposed is $g' = m(s_T)$, where the final state is the new goal. More methods of additional target setting are discussed in [29].

## IV. FORCE-SENSING DEEP DETERMINISTIC POLICY GRADIENT CONTROL

In object pushing tasks, it is found that the training results tend to achieve some abnormal performances (such as oscillation, impact or collision), which are unacceptable. The idea of reasonable improvement is to introduce force observations as a norm for learning behavior.

Our algorithm uses the feature of sparse rewards of the task itself and just discretizes the force-related data (even binarized), since the tactile sensors and force sensors are diffcult to be calibrated. In both the simulation and the real world, we get the data of the torque sensor and the tactile sensor in every step, and use two explanatory functions $I_{ft}$ and $I_{touch}$ as characterization. Its physical meaning includes whether it collides with the desktop or obstacles and whether it is in contact with the block.

The main approaches the force-related data used are:

1. Add two explanatory functions $I_{ft}$ and $I_{touch}$ to the observation space.
2. Reshape sparse reward expression, taking into account $c_i I_i$ ($i \in force\ or\ torque\ sensors$)
3. Add a safe control strategy when exploring the environment.

In order to explore the environment and discover novel and potentially preferable solutions, the sparse reward $r_d(s, g)$ is defined:

$$r_d(s,g) = \begin{cases} 0, & \text{if } g \text{ is satisfied in } s \\ -1, & \text{otherwise} \end{cases} \quad (6)$$

The auxiliary reward of force perception $r_{ft}(s)$ introduced is:

$$r_{ft}(s) = c_{ft} I_{ft}(s), \quad (7)$$

Where

$$I_{ft}(s) = \begin{cases} -1, & \text{if force/torque sensor's amplitude } \geq \sigma_{[ft]} \\ 0, & \text{otherwise} \end{cases} \quad (8)$$

And the auxiliary reward of haptic perception $r_{touch}(s)$ is:

$$r_{touch}(s) = c_{touch} I_{touch}(s) \quad (9)$$

Where

$$I_{touch}(s) = \begin{cases} 1, & \text{if touch sensor's value } \geq \sigma_{[touch]} \\ 0, & \text{otherwise} \end{cases} \quad (10)$$

Algorithm 1 summarizes the training procedure, where $k_{ft}$ represents the safety gain of the auxiliary action, $O_{ft}$ is the observation value of the force/torque sensor, and $\theta$ and $\phi$ are the parameters for the actor and critic network respectively.

**Algorithm 1** Force-based Reinforcement Learning with DDPG and HER
**Require:** Safety gain $k_{ft}$
1: Initialize Actor: $\theta \leftarrow$ random weights
2: Initialize Critic: $\varphi \leftarrow$ random weights
3: **for** episode = 1,N **do**
4:   **for** t=0,T-1 **do**
5:     Add $I_{ft}(s_t)$ and $I_{touch}(s_t)$ in $s_t$
6:     $a_t \leftarrow \pi_\theta(s_t, g) + k_{ft} O_{ft} + \epsilon$, where $\epsilon \sim \mathcal{N}$
7:   **end for**
8:   $\tau \leftarrow (s_0, a_0, ...s_T)$
9:   **for** each $s_t, a_t$ in $\tau$ **do**
10:     $r_t \leftarrow r_d(s_t, g) + r_{ft}(s_t) + r_{touch}(s_t)$
11:   **end for**
12:   Store $(\tau, \{r_t\}, g)$ in replay buffer $R$
13:   Sample episode $(\tau, \{r_t\}, g)$ from $R$
14:   **with** probability $k$
15:     Replay new goal $g'$ with HER
16:     Store $(s_t, s_{t+1}, r'_t, g')$ in $R$
17:   **end with**
18:
19:   **for** each $t$ **do**
20:     $\hat{a}_{t+1} \leftarrow \pi_\theta(s_{t+1}, g)$
21:     $\hat{a}_t \leftarrow \pi_\theta(s_t, g)$
22:     $q_t \leftarrow r_t + \gamma Q_\varphi(s_{t+1}, \hat{a}_{t+1}, g)$
23:     $\Delta q_t \leftarrow q_t - Q_\varphi(s_t, a_t, g)$
24:   **end for**
25:   $\nabla_\varphi = \frac{1}{T} \sum_t \Delta q_t \frac{\partial Q_\varphi(s_t, a_t, g)}{\partial \varphi}$
26:   $\nabla_\theta = \frac{1}{T} \sum_t \frac{\partial Q_\varphi(s_t, \hat{a}_t, g)}{\partial a} \frac{\partial \hat{a}_t}{\partial \theta}$
27:   Update value function and policy with $\nabla_\theta$ and $\nabla_\varphi$
28: **end for**

Apart from the extra action feedback, we also add some safety constraints. First, we set a maximum velocity of end effector beyond 0.1 m/s. Besides, the joint torque (or electric current) is measured in realtime, and the robot will perform a emergency stop when the threshold value is exceeded. Those ways further guarantee the safety of sim-to-real transfer.

## V. EVALUATION EXPERIMENTS

### A. Experimental Setup

A 6-DoF manipulator (Universal Robots, ur5e) is used for object pushing, both in simulation and real-world experiments. In the experiments, the robotic manipulator need to push the cube from an initial position to the target position (within 4 cm). The initial state of manipulator is fixed, where the gripper keeps straight down. The object's initial position and orientation is arbitrary, and the goal is also random (both in manipulator's work space). The object used is a wood block with sides approximately 7 cm. Different from the gym's fetch simulation environment [30], we set the proper soft contact and friction between table and gripper. When the gripper touches the table top, it could still move. But if the contact force is large, the gripper will stick into the table and can't move again. This environmental configuration is more close to reality. Hydraulic touch sensors [31] are mounted on the fingertips, and a 6-axis force torque sensor is installed inside the manipulator. The observation space of agent includes force/torque (ft) data, tactile data, object's 6D pose, the relative pose between object and gripper, the velocity of gripper and the distance between object and goal. Fig.2 demonstrates the experimental setups of simulator and real world, respectively.

### B. Simulation Experiments

All simulations are performed in the MuJoCo physics simulator [32]. The position mode is used for manipulator control, and Gravity is turned on to simulate the reality. Although the virtual touch sensors have been installed on the fingertips, in real world the contact-rich tasks are complicated enough to be simulated. Here we simply set the tactile data as boolean variables. i.e., when the touch force exceeds the threshold limit, we consider that the contact has happened, which ensures the reliability of perception. In simulation, the max time steps of each episode is 200 (20 seconds), and success rate is evaluated from 20 random test episodes. Finally, in order to compare the effects of different algorithms, four environmental rewards are defined as:

- **using force/torque sensor and touch sensors**
  $r_1 = r_d(s,g) + r_{ft}(s) + r_{touch}(s)$
- **only using force/torque sensor**
  $r_2 = r_d(s,g) + r_{ft}(s)$
- **only using touch sensors**
  $r_3 = r_d(s,g) + r_{touch}(s)$
- **without force/torque and touch sensors**
  $r_4 = r_d(s,g)$

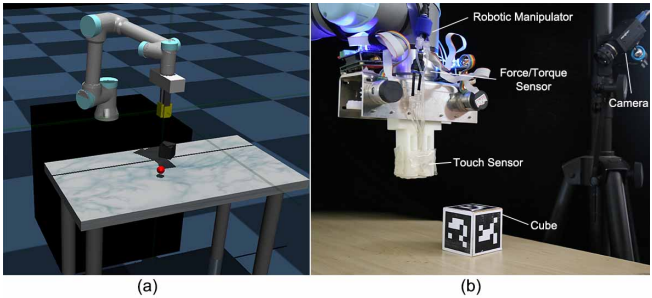

Fig. 2. Experimental setups for object pushing. (a) and (b) represent the simluation environment and real-world environment, respectively.

Table I details the parameters of environment and algorithm.

TABLE I
PARAMETERS OF ENVIRONMENT AND REWARD

| Parameter | Value |
|---|---|
| Force/torque sensor reward gain $c_{ft}$ | 0.2 |
| Force/torque sensor trigger threshold $\sigma_{[ft]}$ | 50 N |
| Touch sensors reward gain $c_{touch}$ | 0.2 |
| Touch sensors trigger threshold $\sigma_{[touch]}$ | 0.1 N |
| Table friction coefficient | 0.03 |

The results of reinforcement learning without force and tactile sensors may lead to abnormally finished tasks. Fig. 3(a) and Fig. 3(b) show the methods of forcing the object to slide to target point with a large force and pushing the object scrolling to the goal, respectively. These methods are feasible in simulation environments, but once transferred to real-world environments, even a slight change of environmental parameters could cause the whole strategy to fail. After tactile reward was applied, this kind of problem has been greatly decreased.

Fig. 4 demonstrates the results in simulation quantitatively. It can be deduced that the widely used method at present with no force feedback has the lowest training efficiency and success rate, and using force/torque or touch sensors can help optimize the learning procedure. The use of the touch sensors causes the robotic manipulator to move toward the block and keep pushing, which has a significant effect on the initial stage of the training. Instead, force/torque sensor avoids the manipulator from dangerous operations, such as excessive output force or potential collisions, which ultimately causes the pushing task to eventually converge to a higher success rate. The experimental environment combining the two sensors has better performance in all stages of training. It can be concluded that the use of force/torque and touch sensors enhances reinforcement learning process in simulation.

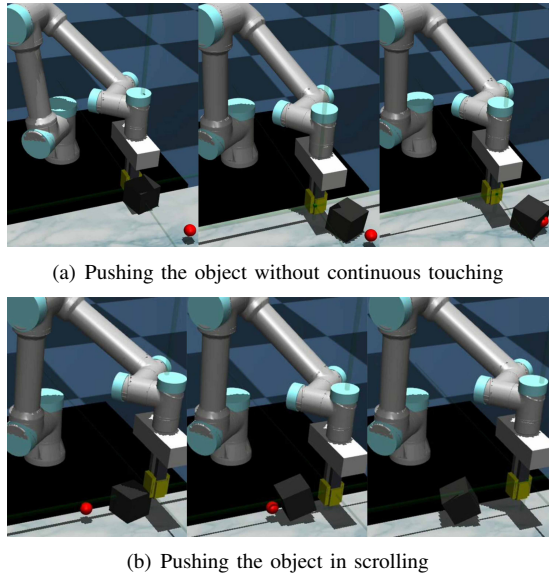

(a) Pushing the object without continuous touching

(b) Pushing the object in scrolling

Fig. 3. Unnatural policy learned by reinforcement learning without force sensing in simulation.

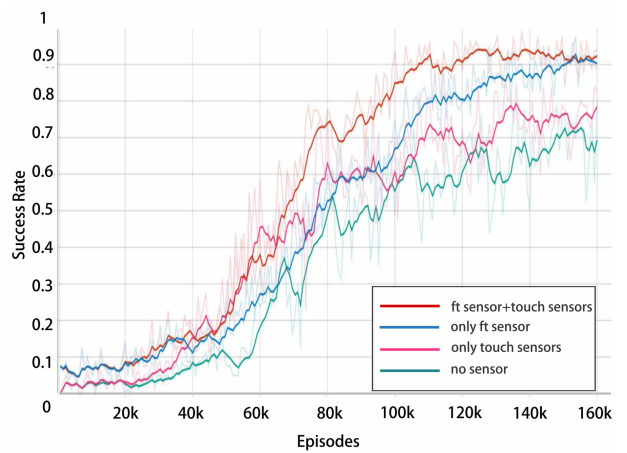

Fig. 4. Learning curves under four strategies.

### C. Real-world Experiments

To display the reliability and safety of our methods, we directly transfer the policy from simulation to real world. A high resolution industrial camera (pointgrey, GS3-U3-28S5C-C, 1920x1440 resolution) is adopted to estimate object's 6D pose using aruco markers [33]. Because of the reality gap and noise of sensors, the transfer could be hard, even if the environmental parameters (such as camera extrinsic parameters, friction coefficient) are calibrated accurately.

In the experiments, inevitably influenced by illumination variations, the random error of pose estimation sometimes exceeds 2 cm, and this error can not be distinguished from object's normal motion. This give rise to the abnormal action of manipulator, illustrated in Fig. 5. When the fingertips get close to the object without touching it, influenced by the variation of observation, the manipulator moves back and forth without pushing the object. Here the employment of touch sensors improves the effect. Based on this fact, when the values of touch sensors are zero (i.e., without touching), the pose of the object should not be changed, we combine the tactile data with machine vision to estimate the object pose more steadily.

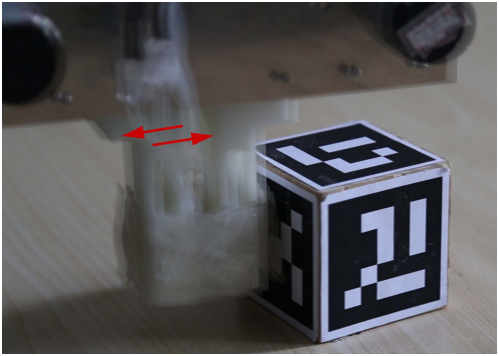

Fig. 5. Abnormal action of manipulator (without tactile perception) in the real world due to visual errors.

In addition, due to the complexity of the strategies obtained from reinforcement learning, the robotic manipulator might collide with the desktop during its motion, which would generate large contact force. In the experiment, our cooperative robot would directly make emergency stop, but most robotic arms with position control mode would damage the environment or break themselves. But our force-sensing safety strategy could well guarantee safety during experiment process.

We performed 20 simulation tests for four different strategies, and 10 times respectively in reality. Fig.6 compares the success rate of object pushing task under different conditions. The experiment not only requires the robotic arm to push the block to the specified position, but also no collision in the whole process, otherwise it will be counted as a failure episode. It can be seen that success rate in real-world environments is far lower than that in simulation, which is mainly caused by domain gap. However, force-based reinforcement learning algorithm improves the success rate greatly in both the simulator and the real world. In addition, the introduction of safety strategy can effectively avoid failures caused by collision. Especially in the sim-to-real transfer process, the force-sensing strategy and the safe control can maintain the higher success rate without prior domain adaptation.

## VI. LIMITATION

Our algorithm combining force-sensing and tactile-sensing improves reliability and efficiency of non-prehensile manipulation to some extent, and guarantees safety in realistic scenes as well. However, there are still deficiencies of our strategy. For instance, visual-based pose estimation plays a very important role in this kind of tasks, but in realistic scenes, circumstances when the visual-based detection doesn't work due to occlusion happen frequently. How to better combine tactile and force sensing for objects' pose estimation will be a major research direction in the future. Furthermore, tactile sensors can be fabricated as an array in order to realize multimodal perception, increase sensing precision and process more complicated contact-rich tasks.

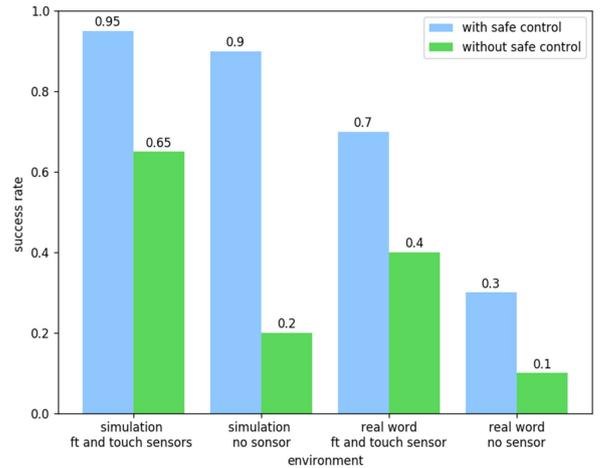

Fig. 6. In simulation and real-world environment, the success rate with force sensing and safe control strategies or not.

## VII. CONCLUSION

In this paper, we proposed an algorithm combining force-sensing, tactile sensing and reinforcement learning to realize safe control and sim-to-real transfer. Our method gives better performances in both simulation and real-world environments. In our future work, we will continue optimizing our policy, and finding better ways to combine force control and reinforcement learning, in order to face more complex and dynamic situations.